\documentclass[letterpaper, 10 pt, conference]{ieeeconf}

\usepackage{graphicx}
\usepackage{algorithm}
\usepackage{algpseudocode}
\algnewcommand{\LeftComment}[1]{\Statex 
\textcolor{gray}{\textit{\(\#\) #1}}}

\usepackage[para]{footmisc}

\usepackage{amsmath} 
\usepackage{amssymb}  
\graphicspath{{figures/}}

\DeclareMathOperator*{\argmax}{arg\,max}

\usepackage{subcaption}

\usepackage{xcolor}
\definecolor{purple}{rgb}{0.54, 0.17, 0.89}

\usepackage{hyperref}
\hypersetup{
    colorlinks=true,
    linkcolor=orange,
    filecolor=magenta,      
    urlcolor=orange,
    citecolor=orange,
}

\usepackage{comment}

\begin{document}
\newcommand{\TODO}[2]{\textcolor{red}{TODO(#1): #2}}
\newcommand{\mkbot}{\textbf{manipulation expert}}
\newcommand{\wipingbot}{\textbf{table wiping expert}}
\newcommand{\rtd}{\emph{RTD}}
\newcommand{\insp}{\emph{Inspection}}
\newcommand{\selfinsp}{\emph{Self-Inspection}}
\newcommand{\planner}{\emph{LLM planner}}
\newcommand{\algname}{VADER}
\newcommand{\algnamelong}{Visual Affordance Detection and Error Recovery}
\newcommand{\pali}{PaLI}
\newcommand{\vild}{ViLD}
\newcommand{\site}{\href{https://google-vader.github.io/}{https://google-vader.github.io/}}
\newcommand{\tweaked}[1]{#1}

\title{\LARGE \bf \algname{}: \algnamelong{}\\for Multi Robot Human Collaboration
}

\author{
Michael~Ahn$^{1}$
\and
Montserrat~Gonzalez~Arenas$^{1}$
\and
Matthew~Bennice$^{2}$
\and
Noah~Brown$^{5}$
\and
Christine~Chan$^{1}$
\and
Byron~David$^{1}$
\and
Anthony~Francis$^{4}$
\and
Gavin~Gonzalez$^{6}$
\and
Rainer~Hessmer$^{2}$
\and
Tomas~Jackson$^{6}$
\and
Nikhil~J~Joshi$^{1}$
\and
Daniel Lam$^{2}$
\and
Tsang-Wei~Edward~Lee$^{1}$
\and
Alex~Luong$^{6}$
\and
Sharath~Maddineni$^{1}$
\and
Harsh~Patel$^{2}$
\and
Jodilyn~Peralta$^{6}$
\and
Jornell~Quiambao$^{5}$
\and
Diego~Reyes$^{5}$
\and
Rosario~M~Jauregui Ruano$^{6}$
\and
Dorsa~Sadigh$^{1}$
\and
Pannag~Sanketi$^{1}$
\and
Leila~Takayama$^{3}$
\and
Pavel~Vodenski$^{2}$
\and
Fei~Xia$^{1}$
\and \\{\hspace{-35pt}\bf(Authors are listed in alphabetical order.)}
}

\maketitle

\renewcommand{\thefootnote}{\arabic{footnote}}
\footnotetext[1]{\href{http://g.co/robotics}{Google DeepMind}}
\footnotetext[2]{\href{https://everydayrobots.com/}{Everyday Robots}}
\footnotetext[3]{\href{https://www.hokulabs.com/}{Hoku Labs}}
\footnotetext[4]{\href{https://www.logicalrobotics.com/}{Logical Robotics}; work begun while at Google Brain Robotics}
\footnotetext[5]{FS Studio}
\footnotetext[6]{Relentless Adrenalin}

\begin{abstract}
Robots today can exploit the rich world knowledge of large language models to chain simple behavioral skills into long-horizon tasks.
However, robots often get interrupted during long-horizon tasks due to primitive skill failures and dynamic environments.
We propose \algname, a \emph{plan, execute, detect} framework with \emph{seeking help} as a new skill that enables robots to recover and complete long-horizon tasks with the help of humans or other robots.
\algname\ leverages visual question answering (VQA) modules to detect \emph{visual affordances} and recognize \emph{execution errors}. It then generates prompts for a language model planner (LMP) which decides when to seek help from another robot or human to \emph{recover from errors} in long-horizon task execution.
\tweaked{
We show the effectiveness of \algname{} with two long-horizon robotic tasks.
Our pilot study showed that \algname{} is capable of performing complex long-horizon tasks by asking for help from another robot to clear a table.
Our user study showed that \algname{} is capable of performing complex long-horizon tasks by asking for help from a human to clear a path. We gathered feedback from people (N=19) about the performance of the \algname{} performance vs. a robot that did not ask for help.
}
\site{}
\end{abstract}

\section{INTRODUCTION}

\begin{figure*}[thpb]
  \centering
  \includegraphics[width=0.7\textwidth]{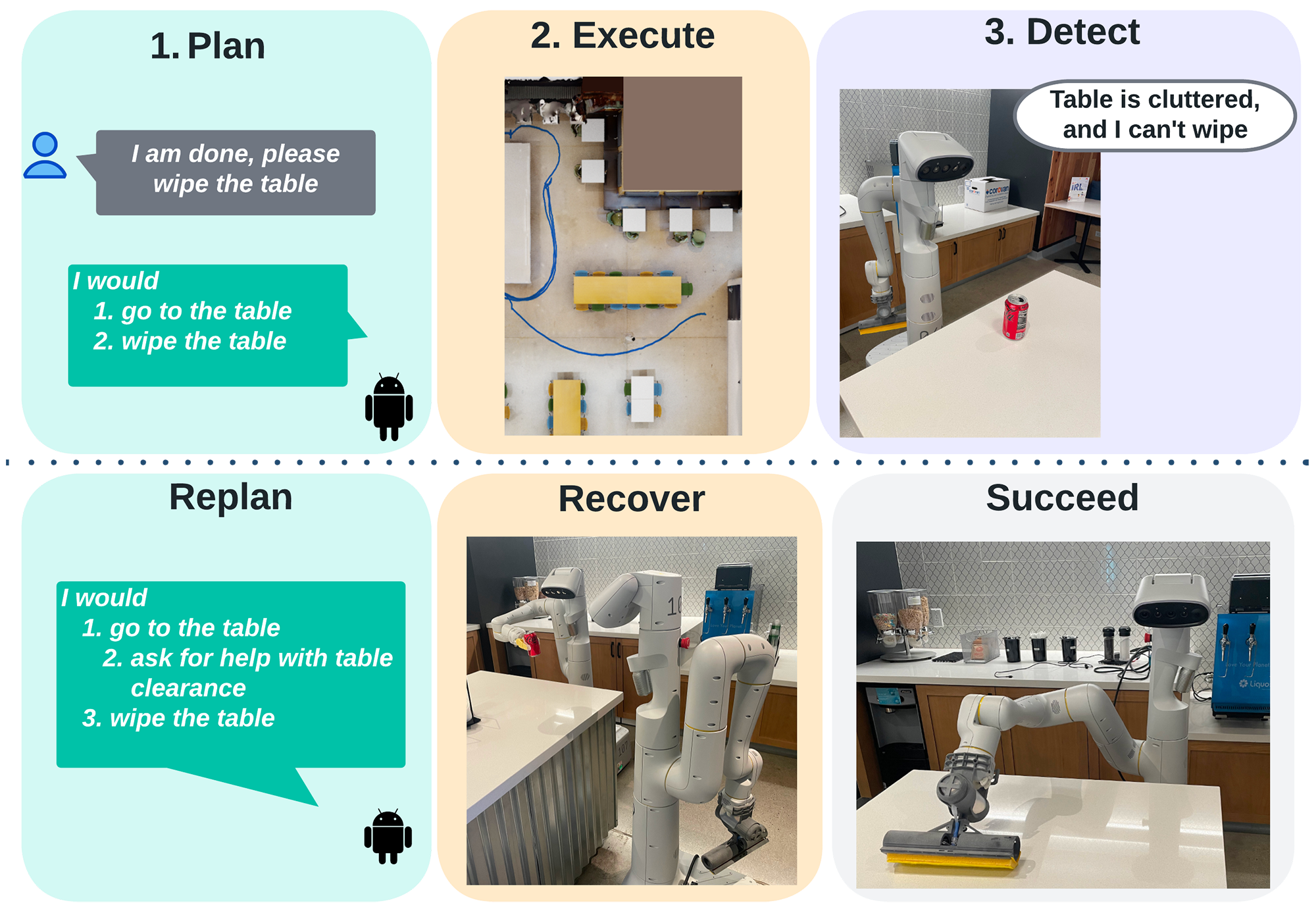}
  \caption{\label{fig:vader} \textbf{\algname{}: \algnamelong{}}. A \emph{plan, execution, detect} framework with a \emph{seeking help} skill as a recovery mechanism. While executing a plan, the robot detects a deviation from its expectations -- a coke can obstructing the table area to be wiped. It replans for recovery, and upon receiving help, completes the original plan.}
\end{figure*}

Task and motion planning is a popular framework for solving long-horizon robotic problems which treats high-level planning and low-level skills as interdependent~\cite{skills2020wang, doi:10.1146/annurev-control-091420-084139}.
Recently, language model planners (LMPs), which use large language models (LLMs) to orchestrate a library of low-level \emph{skill} primitives, have shown promising results in replacing traditional task planners~\cite{saycan2022arxiv, huang2022language}. 
LMPs exploit LLM's rich semantic structure to generate high-level robot plans given diverse human instructions. 
For example, LMP-based robots can execute plans for instructions like \emph{``get me a drink''} or \emph{``I am thirsty, please help me''}, issued in many languages, even if they have not seen this exact instruction in \emph{any} language, by using LLM knowledge to break these tasks into skills and perform them in the correct causal order.

While LMPs alleviate the need for task planning modules, their language plans are often not fully grounded in the robot's environment~\cite{harnad1990symbol}, which can make it hard to evaluate whether a skill should be executed (skill affordances) or whether a skill has succeeded (error recognition). Incorrectly gauging skill affordances can cause planning errors leading to execution of incorrect skills. In addition, skill failures can disrupt task execution. These issues compound as task horizons become longer: ironically, as a robot becomes more capable, it can fail more often!

For example, LMPs such as SayCan \cite{saycan2022arxiv} ground language plans in skill affordances that are based on value functions associated with an RL policy used for executing each skill. While this promotes selecting a plan with skills with high affordances, LMPs often lack awareness of the current state. If a robot breaks its gripper or its workspace is disrupted, many LMPs cannot recognize these dynamic changes and might proceed with an infeasible plan.
Furthermore, while LMPs use affordances as pre-conditions for selecting the {\it next} skill, they generally assume success at previous skill execution lacking the ability to detect errors.

Recent efforts attempt to address these issues by bringing environmental cues, such as scene descriptions into the planning loop. 
For example, extensions of SayCan such as Inner Monologue \cite{huang2022inner} improve reliability of execution by incorporating a variety of environment feedback into the LMP planning loop.
However, while feedback has been shown to be effective, the existing systems typically focus on failures that can be resolved by a single robot.
If the problem cannot be resolved given the robot's own capabilities -- e.g., a robot gripper breaking -- task execution will still fail.

Our key insight is that by grounding with their environments, robots can detect their failures and collaborate with other robots, and humans to course correct through planning. Today's visual question answering (VQA) systems~\cite{chen2022pali} can provide the required grounding mechanism, where natural language summary on a visual observation can be generated within the context of a query. But a multi robot human collaboration is hard due to lack of a mechanism for distributed communication that enables agents to post or claim tasks and provide assistance to each other.
Concretely, our contributions are: 
\begin{itemize}
\item a general-purpose technique called \textit{\algnamelong{}} (\algname{}) which uses feedback from affordance and error detection to generate requests for help from other agents or humans (Fig.~\ref{fig:vader}). This allows the system to dynamically detect failures and employ recovery measures, thus enabling it to complete long horizon tasks. 
\item a cloud-based communication framework to facilitate this assistive collaboration, instantiated with robot agents working alongside humans. 
\item a demonstration of  the effectiveness of our  approach  through a pilot study on a complex, long  horizon task, where  two  robots  with  differing  morphologies (one with a parallel gripper and one with a wiping tool at its end-effector as shown in Fig.~\ref{fig:exp_setup})  have to work  in  collaboration  to  complete, i.e. the task cannot be done by any one of them alone. 
\item \tweaked{
a user study with 19 participants in office kitchen spaces in which \algname{}'s performance at completing a long-horizon task was assessed compared to a control, which did not ask for help.
}
\end{itemize}

\section{BACKGROUND}
\label{s:prelims}

\noindent \textbf {Prior work on Multi Human-Robot Collaboration.}
There is a plethora of work on both multi-robot and human-robot communication and collaboration \cite{shorinwa_distributed_2023_survey, shorinwa_distributed_2023_tutorial, planning2018schwarting}.
Multi-robot coordination often considers task allocation in settings such as collaborative manipulation~\cite{losey2019learning,choudhury2022dynamic}, UAV formation, and multi-agent search and rescue~\cite{du2017distributed, hayat2020multi}. While these works are limited to collaboration between robots, prior work also considers effective human-robot collaboration~\cite{knepper2013ikeabot}. This includes shared autonomy settings that arbitrate or blend human and robot inputs~\cite{nikolaidis2017human}, approaches towards partner modeling~\cite{xie2020learning,nikolaidis2015efficient}, as well as robots asking for help from nearby people~\cite{veloso2015cobots, hayes2014people, srinivasan2016help, saunderson2021robots, nanavati2021modeling}. To the best of our knowledge, prior work does not consider a multi-human-robot collaboration framework that can detect each agent's affordances nor can effectively recover by asking for help from the agent with the appropriate affordances.
The closest to our work is \cite{knepper2015recovering}, which enables a robot to recover from failure by asking for help. However, this work only asks for help from humans, using a classical planner, and a custom request generators. Instead, we use Language Model Planners (LMP) and visual question answering (VQA) to generate ask for help request from any agent; human or robot.

\noindent \textbf{Low-Level Skills.}
\label{s:primitive-skills}
A general robot-environment interaction can be modeled as a Markov Decision Process (MDP) $\mathcal{M} = (\mathcal{S}, \mathcal{A}, P, \mathcal{R}, \gamma)$ defined over the state space $\mathcal{S}$ capturing the environment and robot state, the robot action space $\mathcal{A}$, the transition probability $P: \mathcal{S} \times \mathcal{A} \times \mathcal{S} \rightarrow [0, 1]$, and a reward function $\mathcal{R}:\mathcal{S} \times \mathcal{A} \rightarrow \mathbb{R}$ with a discount factor $\gamma$. Executing a \emph{policy}  $\pi: \mathcal{S} \rightarrow \mathcal{A}$ in an environment results in a trajectory, i.e., a sequence of states, actions, and rewards $\tau = \{(s_0, a_0, r_0), ... ,(s_N, a_N, r_N)\}$. 
Our goal is to find a policy that optimizes its expected discounted reward, or \emph{return} $G = \mathbb{E}_{\tau \sim \pi}[\sum_{k=0}^N \gamma^kr_k]$~\cite{sutton2018reinforcement} over the trajectories that result from following the policy. 
This objective can be achieved via different approaches. For example, in reinforcement learning (RL), the goal is to find such a policy through online interactions with the environment, while in imitation learning (IL) algorithms, e.g., as in behavioral cloning (BC), one assumes access to a set of expert demonstrations that optimize the expected reward, and the goal is to find a policy that matches these expert trajectories.
In model predictive control (MPC), the policy implicitly attempts to achieve high returns by optimizing a proxy \textit{cost function} over forward rollouts of trajectories based on the state-action transition probability $P$.

A \emph{skill} $\sigma$ refers to a sensorimotor primitive described in natural language as defined in~\cite{skills2020wang, saycan2022arxiv}. Every skill $\sigma_i$ has a corresponding policy $\pi_i$. 
In this work, no skill learning was involved: we instead assume access to a library of precomputed policies. In addition, we note that these skills can each be created independently using different methods. 
For example, some of our manipulation policies were trained using BC with a transformer-based architecture~\cite{rt12023arxiv}, while other manipulation policies such as table wiping skill was trained with RL~\cite{tableWiping2022montse}, and our navigation skill uses the non-learned MPC baseline from~\cite{xiao2022learning}.

\noindent \textbf{Language Model Planners.}
Language model planners (LMPs) combine the semantic and causal structure of the world embedded in an LLM with the skills acquired by a robot to construct task execution plans from available skills.
Language model planning involves using a language model to transform a task instruction $\mathcal{T}$ into a plan $\mathcal{P}$ consisting of a sequence of executable skills as defined earlier. For example, the language model can be used to score a fixed set of language representations of available skills $\Sigma$ to produce a ranking for the next skill to be executed $\sigma_n$ in the plan. In~\cite{saycan2022arxiv} the plans were ``grounded''~\cite{harnad1990symbol} by combining the LLM ranking scores of each skill with their corresponding affordances $p^{\mathrm{affordance}}(\sigma|s)$. Specifically using value functions $V(s)$ associated with each skill can act as a proxy for the affordance function 
to obtain what we collectively denote as $p^{\mathrm{LMP}}(\sigma | \mathcal{T}, s, \sigma_{n-1}, ..., \sigma_0)$. 
The skill $\sigma$ that maximizes $p^{\mathrm{LMP}}$ is then selected to be executed next in the task plan. 
We use PaLM~\cite{chowdhery2022palm} as the language model for task planning in our work in a similar fashion as~\cite{saycan2022arxiv}.

\begin{algorithm}[t]
    \caption{\label{alg:main}\algname{}}
    \begin{algorithmic}[1]
        \Require A set of agents $\mathcal{A}^{(j)}$, a set of skills $\Sigma^{(j)}$ with associated execution policies $\Pi^{(j)}$, an outcome description functions $\mathcal{O}_\Sigma$, a task-instruction $\lambda$, and environmental state $s_0$
        \State let $c_n$ be the task execution context update at planning step $n$.
        \State $n = 1, c_0 = \emptyset$
        \While{$c_{n-1} \neq \mathrm{``done"}$}
        \LeftComment{\textbf{1. Plan}: Execution Skill selection (including ``ask for help") using LMP}
            \State $\sigma_n = \underset{\sigma}{\arg\max}\, p^{\mathrm{LMP}}(\sigma | \lambda, s_n, c_{n-1}, ..., c_0)$  
            \LeftComment{Description of successful/failure outcomes of $\pi_\sigma$}
            \State $\ell^\mathrm{exp}_{n+1} = \mathop{\mathcal{O}}(\sigma_n, s_n)$ 
            \LeftComment{\textbf{2. Execute}: the skill by the robot itself or ask for help from an external agent. State evolves to $s_{n+1}$.}
            \State execute $\pi^{(j)}_n(s_n)$ in the environment
            \LeftComment{\textbf{3. Detect}: VQA context affordance }
            \State $\ell^\mathrm{assess}_{n+1} = \underset{\ell}{\arg\max\,} p^{\mathcal{V}_{QA}}(\ell | \ell^\mathrm{exp}_{n+1}, s_{n+1})$ 
            \LeftComment{LMP context update for replanning}
            \State $c_{n} = concat(\sigma_n, \ell^\mathrm{exp}_{n+1},\ell^\mathrm{assess}_{n+1})$ 
            \State $n = n + 1$   
        \EndWhile
    \end{algorithmic}
\end{algorithm}

\noindent \textbf{Visual Question Answering.}
Visual-Language Models (VLM) leverage the abundant (image, text) paired data to learn bi-encoders that map texts and images to the same embedding space $E$ in an attempt to capture semantics and transfer concepts across these two modalities~\cite{radford2021learning, gu2021open, chen2022pali}. VLMs such as CLIP~\cite{radford2021learning} show promising zero-shot classification capabilities to novel concepts based on these encodings, while other VLMs such as \vild~\cite{gu2021open} distill the vision-text knowledge into open vocabulary object detection and mask prediction models. Recently \pali~\cite{chen2022pali} leveraged pretrained LLMs with relatively moderate sized vision models to transfer generalization capabilities acquired by former to the latter. We experimented with all three of these VLM variants in this work, using them as vision question answering (VQA) systems to answer text queries about images with text answers, $\mathcal{V}_{\text{QA}}: \mathcal{I} \times \mathcal{T} \rightarrow \mathcal{T}$.

\begin{figure}[t]
  \centering
  \includegraphics[width=0.4\textwidth]{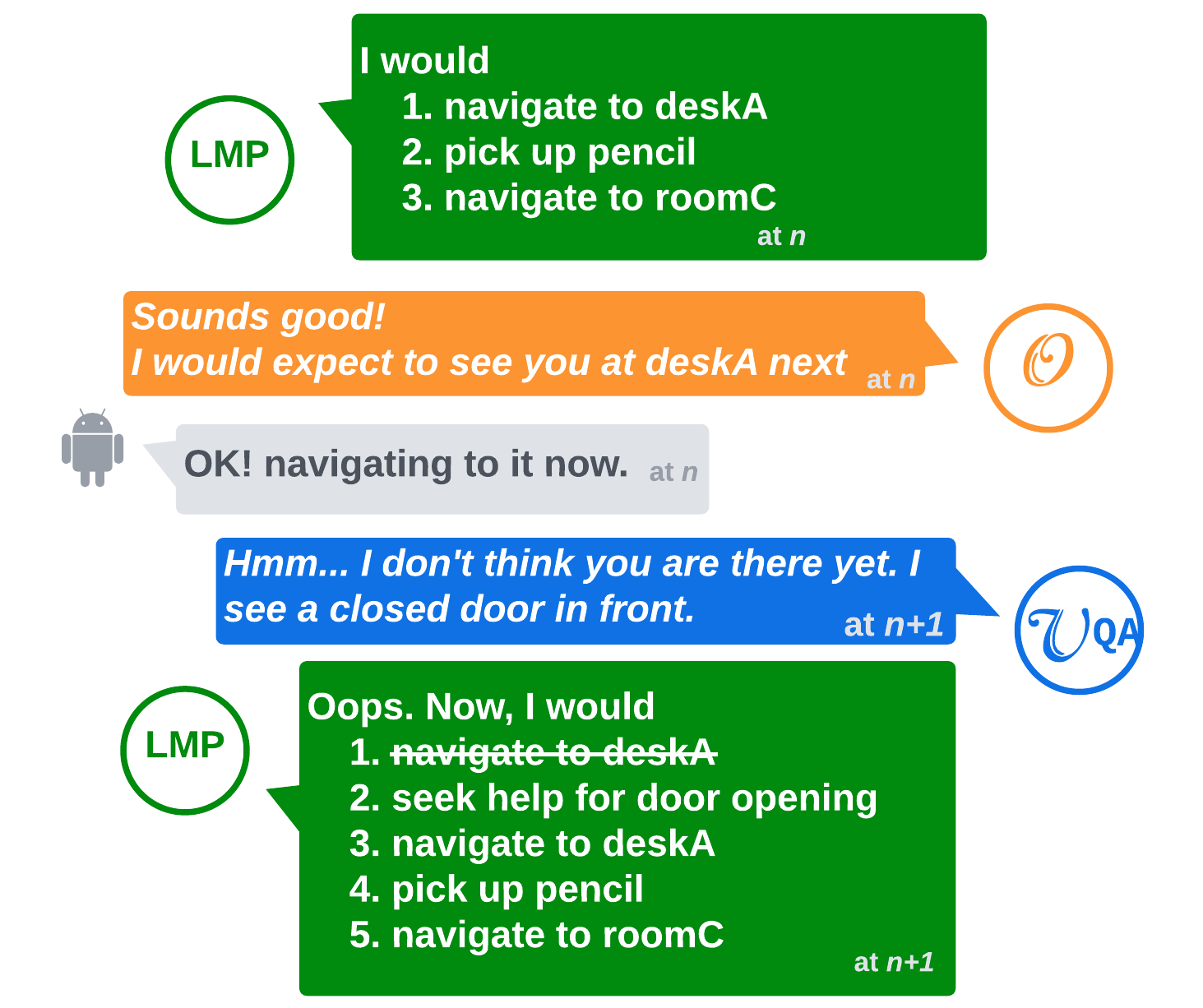}
  \caption{\label{fig:lmp_replan} LMP Replanning with Request for help. The process is imagined as a conversation between different components.
  The immediate next skill from original LMP plan at step $n$ is passed through outcome description function $\mathcal{O}$ and the $\mathcal{V}_{\text{QA}}$ for execution status assessment and folded back to LMP. A recovery plan is laid out. }
\end{figure}

\section{\algname{}}
\label{s:method}

In a nutshell, the problem we are interested in is given a long-horizon natural langauge instruction $\lambda$ such as {\it ``I am done, please wipe the table"} as in Fig.~\ref{fig:vader}, and a library of low-level skills available to a robot $\Sigma$, we would like the robot to generate a task and motion plan that effectively executes the long-horizon task $\lambda$. 

We introduce \algnamelong{} or \algname{}, a \emph{plan, execute, detect} framework, defined in Algorithm~\ref{alg:main}, which uses a VQA to verify the execution of each LMP skill before continuing, similar to the principle of the Test-Operate-Test-Exit loop \cite{miller1960plans} -- but adding the key ability to recover within this loop with the help of other agents. 

The key insight of \algname{} is instead of executing the task plan given the library of skills $\Sigma$ in an open-loop fashion, the robot should detect its visual affordances and perform error recovery by replanning and asking for help from other agents. Specifically \algname{} traverses the loop in Fig.~\ref{fig:vader}: 
\begin{enumerate}
    \item \algname{} first \emph{plans} the next primitive to be executed by selecting the skill $\sigma$ from the library of skills $\Sigma$ with the highest probability for success. For instance, the robot might decide on selecting the skill $\sigma$ = ``pick up coke can''.  We simultaneously generate a language description $\ell^\mathrm{exp}$  = ``coke can in hand'' for the expected desired future state if the execution was successful.
    \item We then \emph{execute} the skill, resulting in a new state $s_{n+1}$ when starting from $s_n$.
    \item We \emph{detect} any deviations or errors by comparing the new state $s_{n+1}$ against the expected description $\ell^\mathrm{exp}$.
\end{enumerate} 

 Deviations could include failures in skill execution, losing the capacity for executing the skill (for example, breaking a gripper), or even incorrectly picking up tasks it can't complete given its skillset. In the case of skill failure, replanning (which occurs during the \emph{plan} and \emph{execute} phases) may bring the robot back to the desired state, as in~\cite{huang2022inner}. In the case of loss of capacity or claiming an out-of-scope task, however, replanning may not be feasible: if unable to complete the task on its own, the agent may need to seek help from another robot or a human to recover (which occurs during the \emph{detect} phase). 
\algname{} thus enables a cooperative environment of robots with diverse skills sets or even morphologies sharing an ecosystem with humans as shown in Fig.~\ref{fig:vader}.

To make this idea concrete, henceforth we assume coexistence of agent variants $\mathcal{A}^{(j)}$'s with diverse skill specializations $\Sigma^{(j)}$. For example, as shown in Fig.~\ref{fig:vader} a robot might have a gripper being able to pick up a coke can while another robot only has a wiping tool at its end-effector allowing it to perform other types of skills such as wiping a table. We assume humans have the ability to perform any of the skills needed for long-horizon tasks. However, to minimally disrupt the autonomy of the overall robotic system, the \algname{} algorithm prioritizes asking for help from other robots before asking for human intervention.

Task execution failures that \algname{} aims to recover from can be grouped broadly according to their causes:

\begin{itemize}
    \item {\bf Infeasible States.} This may be the most common point of failure. For example, a robot may break its gripper, or be blocked on its path, or subscribe to a task requiring a skill outside its capabilities.
    \item {\bf Skill Execution {\it thinks} it was Successful, but Failed}. For example, while a robot is cleaning a table it might drop the debris due to a poor grip, but the manipulation policy may still finish normally. 
    \item {\bf Skill Execution Fails and Alerts.} In rare cases, the skill execution policies are themselves able to sense failures in its execution and halt. For example, a navigation policy may declare infeasible in reach after exhausting all possible paths of approach unsuccessfully.
    \item {\bf Erroneous Planning.} A wrong planning step chosen by LMP can put the execution on an undesirable path eventually making the robot fail. However, because \algname{} relies on the LMP for planning in the loop and has no access to the execution history or broader context, it cannot detect failures in planning.
\end{itemize}

\algname{} adds three key components to an LMP to enable recovery from failures: (a) \textit{detection} of skill affordances and execution errors with visual question answering, (b) \textit{replanning} based on the detected categories of failures, and (c) \textit{recovery} based on seeking help from other agents. 

Note steps (a) and (b) are similar to the closed-loop feedback proposed in~\cite{huang2022inner}, with the key innovations in this work being that in (a) we also check for skill affordance failures where a robot has taken on a skill it either cannot or has become unable to perform; detecting these out-of-scope failures informs the replanning choices in (b) to consider assistance from other agents via (c). Another key difference is that asking for help on a failed \emph{skill} can result in an entirely new \emph{task}, which itself may require the execution of several skills for its completion. For instance, the wiping robot in Fig.~\ref{fig:vader} might realize it can't wipe when there is a coke can on the table. Asking for help will lead to another robot performing a long-horizon task of de-cluttering the table by navigating to it and picking up the coke can.

\subsection{\emph{Detection}: VQA for Affordance Evaluation}
\label{s:affordance}

The main purpose of VQA in \algname{} is to estimate skill affordances in a closed-loop, policy-agnostic way -- as opposed to estimating skill affordances based on RL policy value functions as in~\cite{saycan2022arxiv}, which are mainly used for open-loop skill selection during planning, prior to execution, and may not be trivially available for BC or MPC. From the perspective of \algname{}, we consider value function affordances, if used at all, to be implicitly embedded in $p^\mathrm{LMP}$. 

VQA-based affordance detection can be applied in a plug-and-play fashion using the current state and a language representation of the expected outcomes of the executed skill $\ell^\text{exp}$.
We denote the skill \emph{outcome} description function responsible for generating natural language descriptions of the possible states $s_{n+1}$ given successful or unsuccessful execution of a skill by $\mathcal{O}: \Sigma \times S \rightarrow \Lambda$. In its simplest form, $\mathcal{O}$ could be a lookup table from skill descriptions to outcome descriptions. However, the LMP could be modified to output both the selected skill $\sigma_n$ to be executed at $s_n$ and the expected outcome $\ell^\mathrm{exp}_{n+1}$ at time step $s_{n+1}$. 
If knowledge of the current state is required for this assessment, the same VQA used for estimating skill affordances can be used for assessing the expected outcome $\ell^\mathrm{exp}_{n+1}$ by comparing the skill description $\sigma_n$ with the current state $s_{n}$.

From the perspective of \algname{}, the output of VQA is a language assessment of the last skill execution $\ell^\mathrm{assess}_{n+1}$ which is appended to the skill $\sigma_n$ and fed back to the LMP. In its simplest form, $\ell^\mathrm{assess}_{n+1}$ could be computed over a fixed set of expected outcomes of skill execution over which we pick the answer with the maximum score $\argmax_{\ell} p^{\mathcal{V}_{\text{QA}}}(\ell, s_{n+1})$, as in the case of zero-shot VQA based on \vild{} or CLIP. Alternately, an open-set VQA system like \pali{} could compute $\ell^\mathrm{assess}_{n+1}$ from the current state $s_{n+1}$ based on a text query given the expected outcomes $\ell^\mathrm{exp}_{n+1}$. 

\noindent \textbf{For error detection,} in the navigation example, 
the VQA prompt could be {\it ``is the robot at posA"} applied to an image of the current localization state, and expected outcomes might include ``$\{\text{no}, \text{yes}\}$'' with ``yes'' denoting success. If the answers are scored ``$\{\text{no}: 0.55, \text{yes}: 0.32, ...\}$" then the assessed execution success of ``navigate to destination posA" $\ell^\mathrm{assess}_{n+1}$ would be ``no", resulting on the LMP replanning to recover from failure on the next step.

\noindent \textbf{For affordance detection,} we need to check the precondition of an affordance prior to execution; this is performed by novel \emph{information-gathering skills} that we have created and which the LMP can use to check the preconditions of the next skill. For example, a wiping robot may need the table to be clear of clutter prior to being wiped. Therefore, the LMP may break the task ``\emph{wipe the table}'' into the skills ``\emph{drive to the table}'', ``\emph{check if the table is clear}'', and ``\emph{perform table wiping}''. The skill ``\emph{check if the table is clear}'' is an information gathering skill that checks the prerequisites for  ``\emph{perform table wiping}'' by looking at the table. $\ell^\mathrm{exp}_{n+1}$ for this skill may be ``\emph{is the table clear for wiping?}'' applied to the current camera image using an open-set VQA system like \pali{}. If the answer is ``no'' then the prerequisite for the ``\emph{perform table wiping}'' skill would fail and the LMP would replan by asking another agent to remove the clutter.

\subsection{\emph{Replanning:} Absorbing Failures in the LMP}

The previously selected skill $\sigma_n$ is appended with the outcome expectation $\ell^\mathrm{exp}_{n+1}$ and the execution success assessment $\ell^\mathrm{assess}_{n+1}$ to form a new \emph{context} prompt $c_n := concat(\sigma_n, \ell^\mathrm{exp}_{n+1}, \ell^\mathrm{assess}_{n+1})$ that is fed back to the LMP for replanning.
In the earlier example of the LMP planned skill of {\it ``navigate to destination posA"}, failure would result in the context prompt as  {\it ``navigate to destination posA. at posA? no"}. 
The LMP would generate a new plan as shown in Fig.~\ref{fig:lmp_replan}.

\subsection{\emph{Recovery:} Seeking Help}
For a robot to be a useful, autonomous assistant, it needs ways to recover from failures, preferably without intervention from the original task requester -- but that does \emph{not} mean that the robot cannot request help.
While \algname{} can handle cases where a robot can recover by retrying the same task itself, as in~\cite{huang2022inner}, in many practical scenarios this is neither desirable nor feasible.
Instead, in \algname{} a robot which halts during task execution can request help from a nearby human or from another robot. While we assume humans are so skilled that the environment always affords them completing any task, in order to preserve autonomy of the overall robotic system in aggregation, \algname{} prefers receiving help from another robot before asking a nearby human for help.

\noindent \textbf{Vision Models For Human Detection.}
To seek help from a nearby human, \algname{} relies on the robot's native hardware and software capabilities for human entity and depth perception and does not assume any specific dependency. \algname{} also does not assume human intent prediction capabilities essential for effective social interaction in crowded spaces. In this work, we always pick the nearest human to request help from.
Note, even if a robot cannot detect nearby humans, it can use the Human Robot Fleet Orchestration Service (HRFS) defined in the next section to ask for help.

\noindent \textbf{Human Robot Fleet Orchestration Service.}
\tweaked{
When seeking help from another agent, \algname{} does not assume the agent is physically close. Also, a robot that is currently asking for help from others may later accept a request for help coming from another agent. To facilitate communication between a large fleet of robots and humans we introduce Human Robot Fleet Orchestration Service (HRFS).
}

\tweaked{
HRFS is a real-time transactional communication service supporting multimodal communication among its participants.
HRFS is cloud-based, so it is not running on any specific robot.
HRFS offers an interface where any agent, robot or human, can join the service with a compatible API or app, making it agent agnostic.
After joining, agents can post tasks which can be claimed by other agents. 
For example, a robot may push a task like {\it ``open the door"}, potentially with an executor preference of {\it ``human"}.
HRFS scales well enough to support both the heterogeneous robot fleet in our tests as well as several human operators. 
}

\section{EXPERIMENTS}
\label{s:experiments}

We designed our pilot study and user study to answer the following questions:
\tweaked{
\textbf{1.} the pilot study asked whether \algname{} can enable robots to complete complex, long horizon tasks by detecting failures plus cooperating with each other or humans, 
and \textbf{2.} 
the user study focused on how users reacted to \algname{} compared to a robot that did not ask for help.
}

\subsection{Experiment Setup}
\begin{figure*}[thpb]
  \centering
  \includegraphics[width=\textwidth]{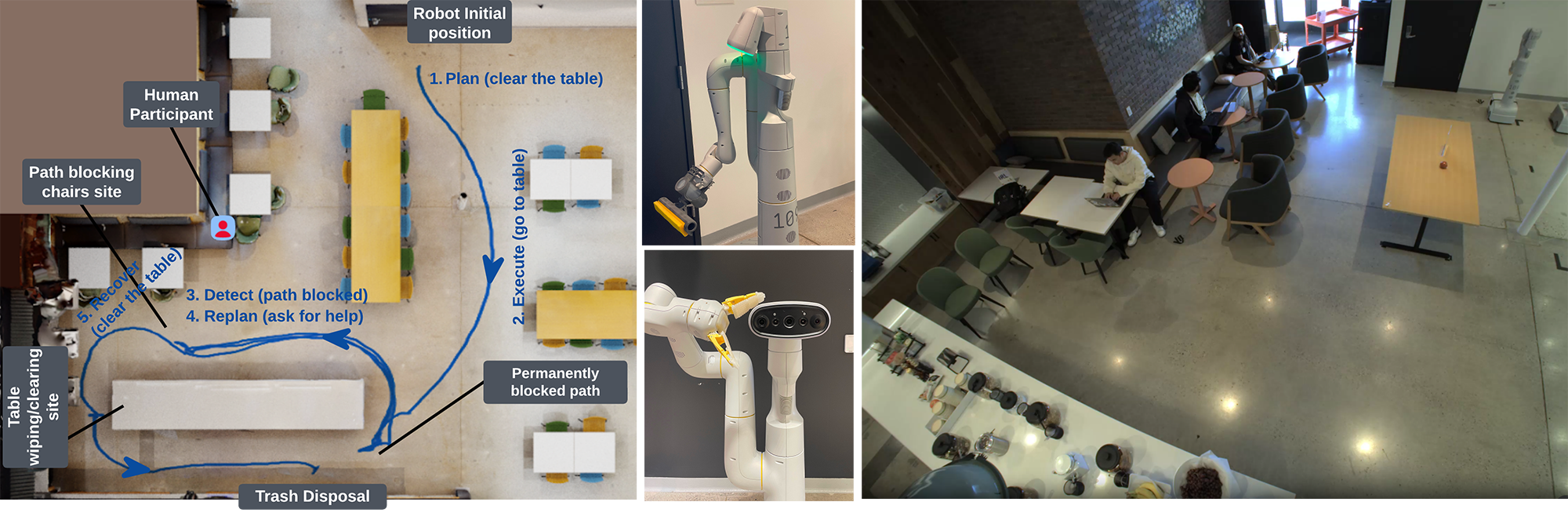}
  \caption{\label{fig:exp_setup} \textbf{Experimental Setup.} \textbf{Left} The experimental setup with robot trajectory overlaid onto the {\it matterport} scanned version of the space used for experiments. {\bf Middle.} (top) \wipingbot{} with the wiping tool, (bottom) \mkbot{} with gripper. \textbf{Right.} A snapshot from one of experiments from our user study discussed in sec.~\ref{s:user-studies}.}
\end{figure*}

We use two kinds of robots in our experiments. The \mkbot{} (ME) has a gripper attached to its arm suitable for executing pick-and-place tasks (Fig.~\ref{fig:exp_setup} middle top), while the the \wipingbot{} (TW), has a specialized tool for wiping table surfaces (Fig.~\ref{fig:exp_setup} middle bottom).

A common office kitchen room with a beverage and snack area, including some chairs and tables steps away from the snack area is used for conducting all the experiments.

We first demonstrate our approach with a complex, long-horizon task that two robots with different morphologies achieve by working in collaboration.  
\tweaked{By design, neither of the robots is capable of finishing the task alone; successful completion necessarily requires multi-robot collaboration.}
In Sec.~\ref{s:user-studies}, we will discuss human-robot collaboration.

We use a setup as shown in Fig.~\ref{fig:exp_setup}. The two robots are parked in the north end of the office kitchen, and are tasked to clean up a table located at the south-west end of the kitchen (annotated as {\it Table wiping site} in Fig.~\ref{fig:exp_setup}). Tasks used were one of {\it ``wipe the table"} -- executable only by the \wipingbot{} -- or {\it ``clear the table"} -- executable by only \mkbot{}. Our system can detect a nearby human~\cite{zhu2020simpose} and request help in human understandable, natural language voice prompts (generated by the LMP). 

Simulating common blockages, both the table wiping site and the navigation routes to the site are blocked with obstacles like a {\it coke can} in former and {\it chairs} in the latter. This enforces collaboration between two robots as well as human intervention to successfully complete the task. 

\subsection{Implementation Details}
\tweaked{
To implement HRFS, we used Firebase Realtime Database (RTD), which provides all the features necessary for the functionality needed for our experiments. We deliberately make our robots unaware of their morphological differences -- access to gripper or wiping tool -- at the initiation. As a result, any task posted to HRFS could be picked up by any of the robots. We expect the robots to detect its tooling capabilities (gripper vs wiping tool) upon a task assignment and update the execution plan accordingly, i.e., either return the task to HRFS without further action \emph{or} proceed with the task execution.}

We use PaLM~\cite{chowdhery2022palm} based setup similar to~\cite{saycan2022arxiv} for LMP.
For $\mathcal{V}_{\text{QA}}$, we tried three state-of-the-art models , namely CLIP~\cite{radford2021learning}, \vild~\cite{gu2021open}, and \pali~\cite{chen2022pali}. They have different trade-offs between inference speed and accuracy. But in our experiments, the difference in performance is minor.
We used simple lookup tables for the outcome function $\mathcal{O}$.

\subsection{Robot-robot Collaboration Pilot Study}
\label{s:robot-collaboration-studies}
We setup a scenario commonly encountered in offices or homes: an obstructed space occluding task execution. The TW was tasked to {\it ``wipe the table"} where the table was deliberately cluttered with obstacles like a {\it coke can} that obstructs the wiping path. \algname{} empowers TW to seek help from ME to clear the table before further wiping.

\tweaked{
To spice up the challenge, our robot embodiments were different but they did not know it and had identical ``brains.'' That is, our robots were not told whether they were TW or ME variants, but both variants had access to the exact same set of skills (i.e. $\Sigma^\mathrm{(TW)} = \Sigma^\mathrm{(ME)}$, where $\Sigma$ included \emph{Table-Wiping}, \emph{Pick-and-Place}, \emph{Navigation}, and \emph{Inspection}, and inspection refers to examining an environment for the presence of objects), even when each could perform only a subset of the skills meaningfully. As a result, upon receiving a task the robots were required to inspect their embodiments, that is, to detect their variety before proceeding with execution, a skill we called \selfinsp{} (examining the robot for the presence of tools or effectors).
}

The success rate of self-variant classification was $90\%$, while that for table clutter detection (one of the information gathering skills discussed in Sec.~\ref{s:affordance}) was $98\%$. The two robots completed the task in 6 out of 11 trials, 4 out of which were due to robot hardware failures and 1 due to clutter detection failures. Inter-robot communication failed once due to an RTD update failure, but it recovered noninvasively after a network re-connection.

A successful episode from this pilot study can be found on \site{}. 
The episode ran 11 minutes end-to-end. The first 3 minutes were spent from TW claiming the wiping task up to requesting help, the next 4 minutes in ME clearing the table while TW waited on an update from ME, and the last 4 in TE completing table-wiping while ME simultaneously put the coke can in the trash. 
\tweaked{Again note that, by design, success on this task required multi-robot collaboration, and the robots were required to self-inspect to detect their own variety.}

\subsection{User Study}
\label{s:user-studies}
While the pilot study confirmed viability of \algname{} with two robots collaborating on completing table-wiping task, to further assess the effectiveness of \algname{}, we conducted a human-robot interaction study to compare how people would respond to each of two different versions of possible robot behaviors -- (1) not asking for help (control condition), and (2) asking for help autonomously (\algname{}).

\tweaked{
We focused our user study on addressing the research question: \textit{would a robot that asks for help be perceived as less capable than one that does not?}
While we might hope people would perceive robots that asked for help as collaborative and useful, it is also possible that robots which ask for help might seem less competent than robots that simply give up on performing the task.
Prior work on humans assisting robots suggests many factors affect human attitudes towards robots that ask for help \cite{srinivasan2016help,baraglia2017efficient}, so the answer to this research question is not clear leading to this user study question.
}

\subsubsection{HRI Study Design}
\tweaked{
To address this research question,  we ran a within-subjects experiment in which we asked each participant to come into an office kitchen area and interact with a robot once for each of the experiment conditions. We counter-balanced the study for order of presentation of the conditions. A total of 19 volunteers participated in our study, but one of them failed to complete the full set of questionnaires so we did not include their data in the statistical analyses.
}

We selected participants neither familiar with our testing robots nor frequent users of robotics systems.
Upon arrival, we explained to the participants the setup and the interaction interface. The robot was given the goal to \emph{``clear the table''} as the high-level task, where the route to the manipulation site was blocked with two chairs (Fig.~\ref{fig:exp_setup} (left)). At the beginning of each round, the participant was asked to stand near the snack area and wait for the robot to move. Depending on the condition (control or \algname{}) the robot may ask the participant for help completing the task.

After each session, we asked the participant to fill out a brief questionnaire, responding on a 7-point Likert scale to:
\begin{enumerate}
    \item ``The robot asked for help in a timely fashion when help was needed'' 
    \item ``I was able to understand when the robot needed help'' 
    \item ``I was able to successfully help the robot'' 
    \item ``The robot was able to continue with the task after me helping with part of the task'' 
    \item ``The robot was successful at accomplishing the task it was asked to do'' 
    \item ``I feel like I can trust the robot to accomplish the task'' 
    \item ``I feel like I can trust the robot to accomplish other similar tasks'' 
    \item ``I would like to collaborate with this robot in the future''
\end{enumerate}

\begin{figure}[t]
    \centering
    \includegraphics[width=0.9\linewidth]{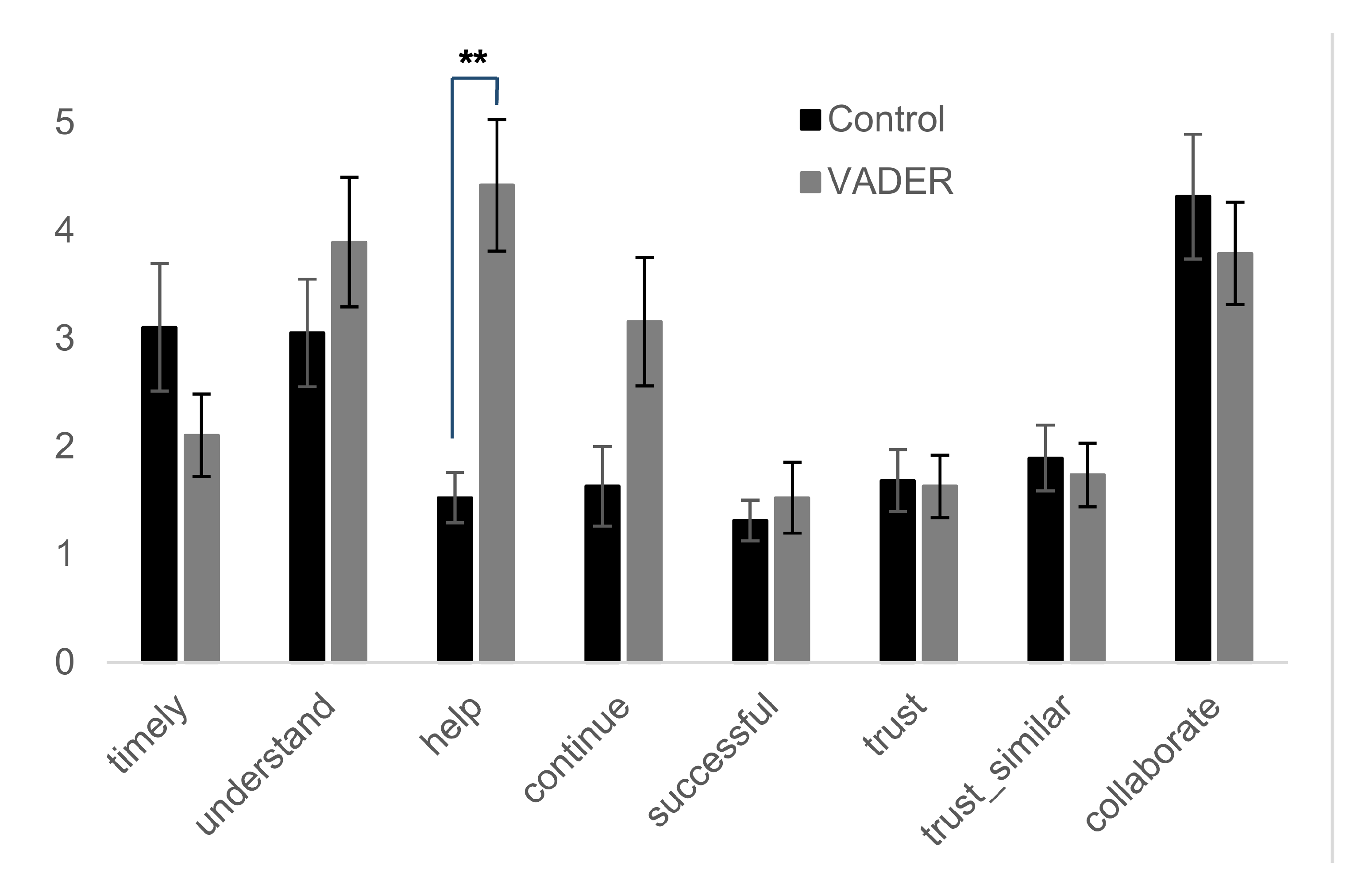}
    \caption{\label{fig:vader_control} Means and standard errors for participant responses to HRI questionnaires. \algname{} outperformed control significantly on the statement where participants were asked if they could help the robot successfully. No significant difference was observed between the  conditions on other statements.}
\end{figure}

\subsubsection{Results}

To evaluate how \algname{} compared to the control condition (the robot not asking for help), we ran a repeated measures analysis of variance (ANOVA) with all levels of the independent variable (\algname{}, control). Because we asked 8 questionnaire items, we used a Bonferroni correction~\cite{econometric2000mittlehammer} to adjust the p-value cut-off for statistical significance, .05 / 8 = .006. 

\tweaked{
Figure~\ref{fig:vader_control} summarizes pairwise contrasts of \algname{} vs. Control. We did not observe statistically significant differences between \algname{} and control on 7 of the 8 statements; however, the two conditions were qualitatively different for statements 1 through 4, which we now discuss.
}

\tweaked{
Participants agreed more strongly with statement 1 (that the robot asked for help in a timely fashion when help was needed) for Control than \algname{}, which is likely due to the greater time that \algname{} took when deciding to ask for help. Conversely, participants agreed more strongly with statement 2 (that they were able to understand when the robot needed help) for \algname{} than Control, which is likely since only \algname{} asked for help.
}

\tweaked{
Participants agreed more strongly with statement 3 (that they were able to help the robot) when it asked for help (running \algname{}, \textit{M}=4.42, \textit{SE}=0.61) than when it did not. This was a significant difference (control condition, \textit{M}=1.53, \textit{SE}=0.23), with pairwise contrast \textit{F}(1,12)=13.78, \textit{p}=.003 $<$ .006. 
We interpret this to mean that users believed they could help the robot when asked. While not a significant difference, participants also agreed with statement 4 (that the robot was able to continue with its task after being helped), which is consistent with this view.
}

\tweaked{
For the remaining statements, \algname{} and Control were neither significantly or qualitatively different.
}

\section{DISCUSSION}


\noindent \textbf{Interpretation.} Our pilot robot collaboration study demonstrated that \algname{} was \textit{viable} as an approach for performing long-horizon robotic tasks using robot-robot cooperation to recover from affordance failures. Our user study expanded the results by showing that people could be incorporated into the \algname{} framework with little to no changes, also enabling recovering from execution errors. While asking for help might be perceived as robot's incompetence, our user study shows that it is not the case with \algname{}. Furthermore, \algname{} significantly improved the success rate 
because it enabled robot-robot and robot-human collaboration.

Robots using \algname{} were always able to complete the task in our user study. However, they scored low on \emph{timely} aspect (Fig.~\ref{fig:vader_control}) compared to the control, which failed fast. 
A preliminary analysis of the task-execution data reveals that our end-to-end episode with \algname{} condition took about 3-4$\times$ longer ($>15$ vs 5 minutes for the control). More than 65$\%$ of the time was spent by the robot in waiting for LMP responses running on cloud (as they could not be deployed on the robot hardware due to their sizes). 
Our hypothesis is that a robot able to recover is useful, but it should act in a timely manner to be desirable as an assistant.

\noindent \textbf{Limitations and Future Work.}
While our work shows that \algname{} enables robots using LMPs to use visual cues to ask for help from robots or humans, the presented approach has a number of limitations, some of which suggest avenues for future work.

\begin{itemize}
    \item \textbf{Study design:} Our study design of users continuously observing the robots until they fail and ask for help does not well represent \algname{}'s target scenario, in which humans are approached only when required. We will design future studies to better align with this scenario.
    \item \tweaked{\textbf{Questionnaire design:} Some of our questionnaire statements were worded in a way which implied that the robot would always ask for help, which did not reflect the control condition. We will refine our questionnaire to more carefully query participants in the future.}
    \item \textbf{Helper determination:} In our study, both humans and robots can respond to tasks. Future work could recommend which agent can better provide help. 
    \item \textbf{Help via dialog:} The LMP uses one voice prompt to request help, but humans may need more context to understand the task. Using dialog-based methods such as Google's Bard~\cite{thoppilan2022lambda} or ChatGPT~\cite{openai2022chatgpt} could improve the likelihood of success. 
    \item \textbf{Respecting social norms:} Our robots search for the nearest person for help. This could be suboptimal: for example, if the nearest person is in conversation, it will be better to ask another person for help. 
\end{itemize}
We are excited about the possibility of building on our work to enable effective human robot collaboration with \algname{} to accelerate progress along the above fronts.

\noindent \textbf{Conclusion.} Large language models have shown the ability to plan over small skills and stitch them together into longer tasks, but this paradoxically has led to increased failure rates due to environment dynamics and skill brittleness. In this paper, we presented \algname{}, an approach which interleaves visual question answering-based error detection and recovery with help of other agents / humans into language model planning, thus allowing a team of humans and robots to achieve complex, long-horizon tasks.

\section*{ACKNOWLEDGMENTS}
The authors would like to thank Carolina Parada, Jie Tan, Ben Jyenis, and Vincent Vanhoucke for their leadership, support and advice on the draft, as well as the robotics operations team at Google for their operators, wranglers and mechatronics engineering support and Everyday Robotics for support.

\bibliographystyle{IEEEtran}
\bibliography{references}
\addtolength{\textheight}{-12cm}   

\end{document}